\documentclass[runningheads]{llncs}
\usepackage{url}
\usepackage{graphicx}
\newcommand{\qheading}[1]{\noindent \textbf{#1}}
\usepackage{amsmath,amssymb} 
\usepackage{color}

\begin{document}
\title{Generating 3D faces using Convolutional Mesh Autoencoders} 

\titlerunning{Generating 3D faces using Convolutional Mesh Autoencoders}
%
\author{Anurag Ranjan \and
Timo Bolkart \and
Soubhik Sanyal \and Michael J. Black}
%
\authorrunning{Anurag Ranjan \and
Timo Bolkart \and
Soubhik Sanyal \and Michael J. Black}
%

\institute{Max Planck Institute for Intelligent Systems \\ T\"ubingen, Germany\\
\email{\{aranjan, tbolkart, ssanyal, black\}@tuebingen.mpg.de}}
\maketitle              
\begin{abstract}
	Learned 3D representations of human faces are useful for computer vision problems such as 3D face tracking and reconstruction from images, as well as graphics applications such as character generation and animation.
Traditional models learn a latent representation of a face using linear subspaces or higher-order tensor generalizations.
Due to this linearity, they can not capture extreme deformations and non-linear expressions.
%
%
To address this, we introduce a versatile model that learns a non-linear representation of a face using spectral convolutions on a mesh surface. 
We introduce mesh sampling operations that enable a hierarchical mesh representation that captures non-linear variations in shape and expression at multiple scales within the model.
In a variational setting, our model samples diverse realistic 3D faces from a multivariate Gaussian distribution.
%
%
Our training data consists of 20,466 meshes of extreme expressions captured over 12 different subjects. 
Despite limited training data, our trained model outperforms state-of-the-art face models with 50\% lower reconstruction error,  while using 75\% fewer parameters. 
We show that, replacing the expression space of an existing state-of-the-art face model with our model, achieves a lower reconstruction error. 
Our data, model and code are available at \url{http://coma.is.tue.mpg.de/}.

\end{abstract}
\section{Introduction}

The human face is highly variable in shape as it is affected by many factors such as age, sex, ethnicity, etc., and deforms significantly with expressions. The existing state of the art 3D face representations mostly use linear transformations \cite{Tewari2017,FLAME2017,Thies2015} or higher-order tensor generalizations \cite{Vlasic2005,Brunton2014,Cao2014_FaceWarehouse}. These 3D face models have several applications including face recognition \cite{taigman2014deepface}, generating and animating faces \cite{FLAME2017} and monocular 3D face reconstruction \cite{tal2017extreme}. Since these models are linear, they do not capture the non-linear deformations due to extreme facial expressions. These expressions are crucial to capture the realism of a 3D face. 




Meanwhile, convolutional neural networks (CNNs) have 
emerged as rich models for generating images \cite{gan,oord2016pixel}, audio \cite{WaveNet_2016}, etc. One of the reasons for their success is attributed to the multi-scale hierarchical structure of CNNs that allows them to learn translational-invariant localized features.
Recent works have explored volumetric convolutions \cite{brock2016generative} for 3D representations. However, volumetric operations require a lot of memory and have been limited to low resolution 3D volumes. Modeling convolutions on 3D meshes  can be memory efficient and allows for processing high resolution 3D structures.
However, CNNs have mostly been successful in Euclidean domains with grid-based structured data and the generalization of CNNs to meshes is not trivial. Extending CNNs to graph structures and meshes has only recently drawn significant attention \cite{Bruna2013,Defferrard2016,Bronstein2017}. Hierarchical operations in CNNs such as max-pooling and upsampling have not been adapted to meshes.
Moreover, training CNNs on 3D facial data is challenging due to the limited size of current 3D datasets. 
Existing large scale datasets \cite{Cao2014_FaceWarehouse,Cosker2011,BU-3DFE_2006,BU-4DFE_2008,Bosphorus_2008} do not contain high resolution extreme facial expressions.

To address these problems, we introduce a Convolutional Mesh Autoencoder (CoMA) with novel mesh sampling operations, which preserve the topological structure of the mesh features at different scales in a neural network. We follow the work of Defferrard et al.~\cite{Defferrard2016} on generalizing the convolution on graphs using fast Chebyshev filters, and use their formulation for convolving over our facial mesh. We perform spectral decomposition of meshes and apply convolutions directly in frequency space. This makes convolutions memory efficient and feasible to process high resolution meshes. We combine the convolutions and sampling operations to construct our model in the form of a Convolutional Mesh Autoencoder. We show that CoMA performs much better than state of the art face models at capturing highly non-linear extreme facial expressions with fewer model parameters. Having fewer parameters in our model makes it more compact, and easier to train. This reduction in parameters is attributed to the locally invariant convolutional filters that can be shared over the mesh surface. 


We address the problem of data limitation by capturing 20,466 high resolution meshes with extreme facial expressions in a multi-camera active stereo system. Our dataset spans 12 subjects performing 12 different expressions. The expressions are chosen to be complex and asymmetric, with significant deformation in the facial tissue.

In summary, our work introduces a representation that models variations on the mesh surface using a hierarchical multi-scale approach and can generalize to other 3D mesh processing applications. Our main contributions are: 1) we introduce a Convolutional Mesh Autoencoder consisting of mesh downsampling and mesh upsampling layers with fast localized convolutional filters defined on the mesh surface; 2) we show that our model accurately represents 3D faces in a low-dimensional latent space performing 50\% better than a PCA model that is used in state of the art face models such as ~\cite{Tewari2017,FLAME2017,Amberg2008,Breidt2011,Yang2011}; 3) our autoencoder uses up to 75\% fewer parameters than linear PCA models, while being more accurate in terms of reconstruction error; 4) we show that replacing the expression space of a state of the art face model, FLAME \cite{FLAME2017}, by CoMA improves its reconstruction accuracy;
5) we show that our model can be used in a variational setting to sample a diversity of facial meshes from a known Gaussian distribution;
6) we provide 20,466 frames of complex 3D head meshes from 12 different subjects for a range of extreme facial expressions along with our code and trained models for research purposes.


\section{Related work}

\qheading{Face Representations.} Blanz and Vetter~\cite{BlanzVetter1999} introduced the \textit{morphable model}; the first generic representation for 3D faces based on principal component analysis (PCA) to describe facial shape and texture variations. We also refer the reader to Brunton et al.~\cite{Brunton2014_Review} for a comprehensive overview of 3D face representations. To date, the Basel Face Model (BFM)~\cite{BFM2009}, i.e. the publicly available variant of the morphable model, is the most widely used representation for 3D face shape in a neutral expression. Booth et al.~\cite{Booth2017} recently proposed another linear neutral expression 3D face model learned from almost $10,000$ face scans of more diverse subjects.

Representing facial expressions with linear spaces, or higher-order generalizations thereof, remains the state-of-the-art. The linear expression basis vectors are either computed using PCA ~\cite{Amberg2008,Breidt2011,FLAME2017,Tewari2017,Yang2011}, or are manually defined using linear blendshapes (e.g.~\cite{Thies2015,Li2010,Bouaziz2013}). Yang et al.~\cite{Yang2011} use multiple PCA models, one per expression, Amberg et al.~\cite{Amberg2008} combine a neutral shape PCA model with a PCA model on the expression residuals from the neutral shape. A similar model with an additional albedo model was used within the Face2Face framework~\cite{Thies2016}. The recently published FLAME model~\cite{FLAME2017} additionally models head rotation, and yaw motion with linear blendskinning and achieves state-of-the-art results. Vlasic et al.~\cite{Vlasic2005} introduce multilinear models, i.e., a higher-order generalization of PCA to model expressive 3D faces. Recently, Fern{\'a}ndez et al.~\cite{Fernandez2018} propose an autoencoder with a CNN-based encoder and a multilinear model as a decoder. Opposed to our mesh autoencoder, their encoder operates on depth images rather than directly on meshes. For all these methods, the model parameters globally influence the shape; i.e.~each parameter affects all the vertices of the face mesh. Our convolutional mesh autoencoder however models localized variations due to the hierarchical multiscale nature of the convolutions combined with the down- and up-sampling.

To capture localized facial details, Neumann et al.~\cite{Neumann2013} and Ferrari et al.~\cite{Ferrari2015} use sparse linear models. Brunton et al.~\cite{Brunton2014} use a hierarchical multiscale approach by computing localized multilinear models on wavelet coefficients. While Brunton et al.~\cite{Brunton2014} also used a hierarchical multi-scale representation, their method does not use shared parameters across the entire domain. Note that sampling in localized low-dimensional spaces \cite{Brunton2014} is difficult due to the locality of the facial features; combinations of localized facial features are unlikely to form plausible global face shapes. One goal of our work is to generate new face meshes by sampling the latent space, thus we design our autoencoder to use a single low-dimensional latent space.

Jackson et al.~\cite{Jackson2017} use a volumetric face representation in their CNN-based framework. In contrast to existing face representation methods, our mesh autoencoder uses convolutional layers to represent faces with significantly fewer parameters. Since it is defined completely on the mesh space, we do not have memory constraints which affect volumetric convolutional methods for representing 3D models.

\qheading{Convolutional Networks.} Bronstein et al.~\cite{Bronstein2017} give a comprehensive overview of generalizations of CNNs on non-Euclidean domains, including meshes and graphs. Masci et al.~\cite{Masci2015} define the first mesh convolutions by locally parameterizing the surface around each point using geodesic polar coordinates, and defining convolutions on the resulting angular bins. In a follow-up work, Boscaini et al.~\cite{Boscaini2016} parametrize local intrinsic patches around each point using anisotropic heat kernels. Monti et al.~\cite{Monti2017} introduce $d$-dimensional pseudo-coordinates that define a local system around each point with weight functions. This method resembles the intrinsic mesh convolution of \cite{Masci2015} and \cite{Boscaini2016} for specific choices of the weight functions. In contrast, Monti el al.~\cite{Monti2017} use Gaussian kernels with a trainable mean vector and covariance matrix as weight functions.

Verma et al.~\cite{Verma2017} presente dynamic filtering on graphs where the filter weights depend on the inputs. This work does not focus on reducing the dimensionality of graphs or meshes. Yi et al.~\cite{Yi2017} also present a spectral CNN for labeling nodes but does not involve any mesh dimensionality reduction. Sinha et al.~\cite{Sinha2016} and Maron et al.~\cite{Maron2017} embed mesh surfaces into planar images to apply conventional CNNs. Sinha et al.~use a robust spherical parametrization to project the surface onto an octahedron, which is then cut and unfolded to form a square image. Maron et al.~\cite{Maron2017} introduce a conformal mapping from the mesh surface into a flat torus. Litani et al.\cite{litany2017deformable} use graph convolutions for shape completion.

Although, the above methods presented generalizations of convolutions on meshes, they do not use a structure to reduce the meshes to a low dimensional space. Our proposed autoencoder efficiently handles these problems by combining the mesh convolutions with efficient mesh-downsampling and mesh-upsampling operators.

Bruna et al.~\cite{Bruna2013} propose the first generalization of CNNs on graphs by exploiting the connection of the graph Laplacian and the Fourier basis (see Section~\ref{sec:mesh_operators} for more details). This leads to spectral filters that generalize graph convolutions. Boscaini et al.~\cite{Boscaini2015} extend this using a windowed Fourier transform to localize in frequency space. Henaff et al.~\cite{Henaff2015} build upon the work of Bruna et al. by adding a procedure to estimate the structure of the graph. To reduce the computational complexity of the spectral graph convolutions, Defferrard et al.~\cite{Defferrard2016} approximate the spectral filters by truncated Chebyshev poynomials, which avoids explicitly computing the Laplacian eigenvectors, and introduce an efficient pooling operator for graphs. Kipf and Welling~\cite{KipfWelling2016} simplify this using only first-order Chebyshev polynomials.

However, these graph CNNs are not directly applied to 3D meshes. CoMA uses truncated Chebyshev polynomials \cite{Defferrard2016} as mesh convolutions. In addition, we define mesh down-sampling and up-sampling layers to obtain a complete mesh autoencoder structure to represent highly complex 3D faces, obtaining state of the art results in 3D face modeling.

\section{Mesh Operators}
\label{sec:mesh_operators}
We define a 3D facial mesh as a set of vertices and edges, $\mathcal{F}=(\mathcal{V}, A)$, with $|\mathcal{V}| = n$ vertices that lie in 3D Euclidean space, $\mathcal{V} \in \mathbb{R}^{n \times 3}$.
The sparse adjacency matrix $A \in \{0,1\}^{n \times n}$ represents the edge connections, where $A_{ij} = 1$ denotes an edge connecting vertices $i$ and $j$, and $A_{ij} = 0$ otherwise. The non-normalized graph Laplacian \cite{chung1997spectral} is defined as $L = D - A$, with the diagonal matrix $D$ that represents the degree of each vertex in $\mathcal{V}$ as $D_{ii} = \sum_{j} A_{ij}$. 

The Laplacian is diagonalized by the Fourier basis $U \in \mathbb{R}^{n \times n}$ (as $L$ is a real symmetric matrix) as $L = U \Lambda U^T$, where the columns of $U = [u_0, u_1, ..., u_{n-1}]$ are the orthogonal eigenvectors of $L$, and $\Lambda = diag([\lambda_0, \lambda_1, ...,$ $ \lambda_{n-1}]) \in \mathbb{R}^{n \times n}$ is a diagonal matrix with the associated real, non-negative eigenvalues. The graph Fourier transform \cite{chung1997spectral} of the mesh vertices $x \in \mathbb{R}^{n \times 3}$ is then defined as $x_\omega = U^Tx$, and the inverse Fourier transform as $x = Ux_\omega$.

\subsection{Fast spectral convolutions} The convolution operator $*$ can be defined in Fourier space as a Hadamard product, $x * y = U((U^Tx)\odot(U^Ty))$. This is computationally expensive with large numbers of vertices, since $U$ is not sparse. The problem is addressed by formulating mesh filtering with a kernel $g_\theta$ using a recursive Chebyshev polynomial \cite{Defferrard2016,Hammond2011}. The filter $g_\theta$ is parametrized as a Chebyshev polynomial of order $K$ given by
\begin{equation}
	g_\theta(L) = \sum_{k=0}^{K-1} \theta_k T_k(\tilde{L}),
\end{equation}
where $\tilde{L} = 2 L/\lambda_{max} - I_n$ is the scaled Laplacian, the parameter $\theta \in \mathbb{R}^K$ is a vector of Chebyshev coefficients, and $T_k \in \mathbb{R}^{n \times n}$ is the Chebyshev polynomial of order $k$ that can be computed recursively as $T_k(x) = 2xT_{k-1}(x) - T_{k-2}(x)$ with $T_0=1$ and $T_1=x$. The spectral convolution can then be defined as in \cite{Defferrard2016}
\begin{equation}
y_j = \sum_{i=1}^{F_{in}} g_{\theta_{i,j}}(L) x_{i} \in \mathbb{R}^n,
\end{equation}
where $y_j$ computes the $j^{th}$ feature of $y \in \mathbb{R}^{n \times F_{out}}$. The input $x \in \mathbb{R}^{n \times F_{in}}$ has $F_{in}$ features. The input face mesh has $F_{in}=3$ features corresponding to its 3D vertex positions. Each convolutional layer has $F_{in} \times F_{out}$ vectors of Chebyshev coefficients, $\theta_{i,j} \in \mathbb{R}^K$, as trainable parameters.


\subsection{Mesh Sampling}
\begin{figure}[t]
\begin{center}
\includegraphics[width=\linewidth]{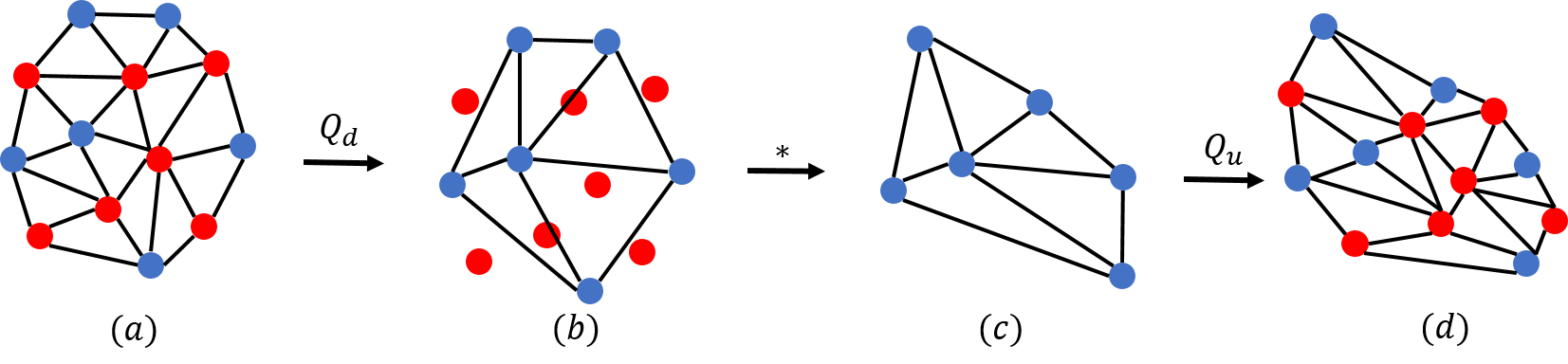}
\end{center}
\caption{Mesh Sampling Operations: A mesh feature (a) is down-sampled by removing red vertices that minimize quadric error \cite{qslim}. We store the barycentric coordinates of the red vertices w.r.t.~the down-sampled mesh (b). The down-sampled mesh can then be transformed using convolutional operations to obtain the transformed mesh (c). The contracted vertices are then added at the barycentric locations (d). }
\label{fig:mesh_sampling}
\end{figure}

In order to capture both global and local context, we seek a hierarchical multi-scale representation of the mesh. This allows convolutional kernels to capture local context in the shallow layers and global context in the deeper layers of the network. In order to address this representation problem, we introduce mesh sampling operators that define the down-sampling and up-sampling of a mesh feature in a neural network. A mesh feature with $n$ vertices can be represented using a $n \times F$ tensor, where $F$ is the dimensionality of each vertex. A 3D mesh is represented with $F=3$. However, applying convolutions to the mesh can result in features with different dimensionality. The mesh sampling operations define a new topological structure at each layer and maintain the context on neighborhood vertices. We now describe our sampling method with an overview as shown in Figure \ref{fig:mesh_sampling}.

We perform the in-network down-sampling of a mesh with $m$ vertices using transform matrices $Q_d \in \{0,1\}^{n \times m}$, and up-sampling using $Q_u \in \mathbb{R}^{m \times n}$ where $m>n$.
The down-sampling is obtained by contracting vertex pairs iteratively that maintain surface error approximations using quadric matrices \cite{qslim}. In Figure~\ref{fig:mesh_sampling}(a), the red vertices are contracted during the down-sampling operation. The (blue) vertices after down-sampling are a subset of the original mesh vertices $\mathcal{V}_d \subset \mathcal{V}$. Each weight $Q_d(p,q) \in \{0,1\}$ denotes whether the $q$-th vertex is kept during down-sampling, $Q_d(p,q) = 1$, or discarded where $Q_d(p,q) = 0,$ $\forall{p}$. 

Since a loss-less down-sampling and up-sampling is not feasible for general surfaces, the up-sampling matrix is built during down-sampling. Vertices retained during down-sampling (blue) undergo convolutional transformations, see Figure~\ref{fig:mesh_sampling}(c). These (blue) vertices are retained during up-sampling $Q_u(q,p) = 1$ iff $Q_d(p,q) = 1.$
Vertices $v_q \in \mathcal{V}$ discarded during down-sampling (red vertices) where $Q_d(p,q) = 0 $ $\forall{p}$, are mapped into the down-sampled mesh surface using barycentric coordinates. As shown in Figures \ref{fig:mesh_sampling}(b)-\ref{fig:mesh_sampling}(d), this is done by projecting $v_q$ into the closest triangle $(i,j,k)$ in the down-sampled mesh, denoted by $\widetilde{v}_p$, and computing the barycentric coordinates, $\widetilde{v}_p = w_i v_i + w_j v_j + w_k v_k$, such that $v_i, v_j, v_k \in \mathcal{V}_d$ and $w_i + w_j + w_k =1$. The weights are then updated in $Q_u$ as $Q_u(q,i) = w_i$, $Q_u(q,j) = w_j$, and $Q_u(q,k) = w_k$, and $Q_u(q,l) = 0$ otherwise. The up-sampled mesh with vertices $\mathcal{V}_u$ is obtained using sparse matrix multiplication, $\mathcal{V}_u = Q_u \mathcal{V}_d$.

\section{Mesh Autoencoder}

\begin{figure}[t]
\begin{center}
\includegraphics[width=\linewidth]{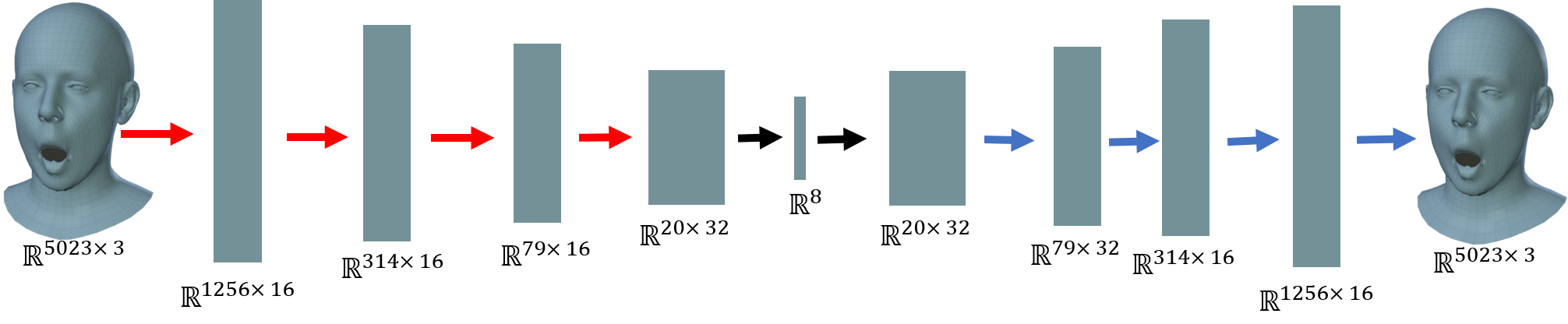}
\end{center}
\caption{Convolutional Mesh Autoencoder: The red and blue arrows indicate down-sampling and up-sampling layers respectively.}
\label{fig:meshae}
\end{figure}

\begin{table}[b]
	\begin{minipage}{.45\linewidth}
		\centering
		\caption{Encoder architecture}
		\begin{tabular}{lcc}
			\multicolumn{1}{l}{Layer}  &\multicolumn{1}{c}{Input Size} &\multicolumn{1}{c}{Output Size}
			\\ \hline
			Convolution         &$5023 \times 3$ &$5023 \times 16$ \\
			Down-sampling             &$5023 \times 16$ &$1256 \times 16$ \\
			Convolution             &$1256 \times 16$  &$1256 \times 16$ \\
			Down-Sampling             &$1256 \times 16$ &$314 \times 16$ \\
			Convolution         &$314 \times 16$ &$314 \times 16$ \\
			Down-Sampling             &$314 \times 16$ &$79 \times 16$ \\
			Convolution             &$79 \times 16$  &$79 \times 32$\\
			Down-Sampling             &$79 \times 32$ &$20 \times 32$ \\
			Fully Connected         &$20 \times 32$ &$8$ \\
		\end{tabular}
		\label{tab:encoder}
	\end{minipage}%
	\hfil
	\begin{minipage}{.45\linewidth}
		\centering
		\caption{Decoder architecture}
		\begin{tabular}{lcc}
			\multicolumn{1}{l}{Layer}  &\multicolumn{1}{c}{Input Size} &\multicolumn{1}{c}{Output Size}
			\\ \hline
			Fully Connected         &$8$ &$20 \times 32$  \\
			Up-Sampling             &$20 \times 32$ &$79 \times 32$  \\
			Convolution             &$79 \times 32$  &$79 \times 32$ \\
			Up-Sampling             &$79 \times 32$ &$314 \times 32$  \\
			Convolution         &$314 \times 32$ &$314 \times 16$ \\
			Up-Sampling             &$314 \times 16$ &$1256 \times 16$ \\
			Convolution             &$1256 \times 16$  &$1256 \times 16$\\
			Up-Sampling             &$1256 \times 16$ &$5023 \times 16$ \\
			Convolution         &$5023 \times 16$ &$5023 \times 3$ \\
		\end{tabular}
		\label{tab:decoder}
	\end{minipage}
\end{table}

\qheading{Network Architecture.} Our autoencoder consists of an encoder and a decoder. The structure of the encoder is shown in Table \ref{tab:encoder}. The encoder consists of 4 Chebyshev convolutional filters with $K=6$ Chebyshev polynomials. Each of the convolutions is followed by a biased ReLU~\cite{Glorot2011}. The down-sampling layers are interleaved between convolutional layers. Each of the down-sampling layers reduce the number of mesh vertices by approximately 4 times. The encoder transforms the face mesh from $\mathbb{R}^{n \times 3}$ to an 8 dimensional latent vector using a fully connected layer at the end.

The structure of the decoder is shown in Table \ref{tab:decoder}. The decoder similarly consists of a fully connected layer that transforms the latent vector from $\mathbb{R}^8$ to $\mathbb{R}^{20 \times 32}$ that can be further up-sampled to reconstruct the mesh. Following the decoder's fully connected layer, 4 convolutional layers with interleaved up-sampling layers generate a 3D mesh in $\mathbb{R}^{5023 \times 3}$. Each of the convolutions is followed by a biased ReLU similar to the encoder network. Each up-sampling layer increases the numbers of vertices by approximately 4 times. Figure \ref{fig:meshae} shows the complete structure of our mesh autoencoder.

\qheading{Training Details.} We train our autoencoder for 300 epochs with a learning rate of 8e-3 and a learning rate decay of 0.99 every epoch. We use stochastic gradient descent with a momentum of 0.9 to optimize the L1 loss between predicted mesh vertices and the ground truth samples. We use L1 regularization on the weights of the network using weight decay of 5e-4. The convolutions use Chebyshev filtering with $K=6$.

\begin{figure}[t]
\begin{center}
\includegraphics[width=\linewidth]{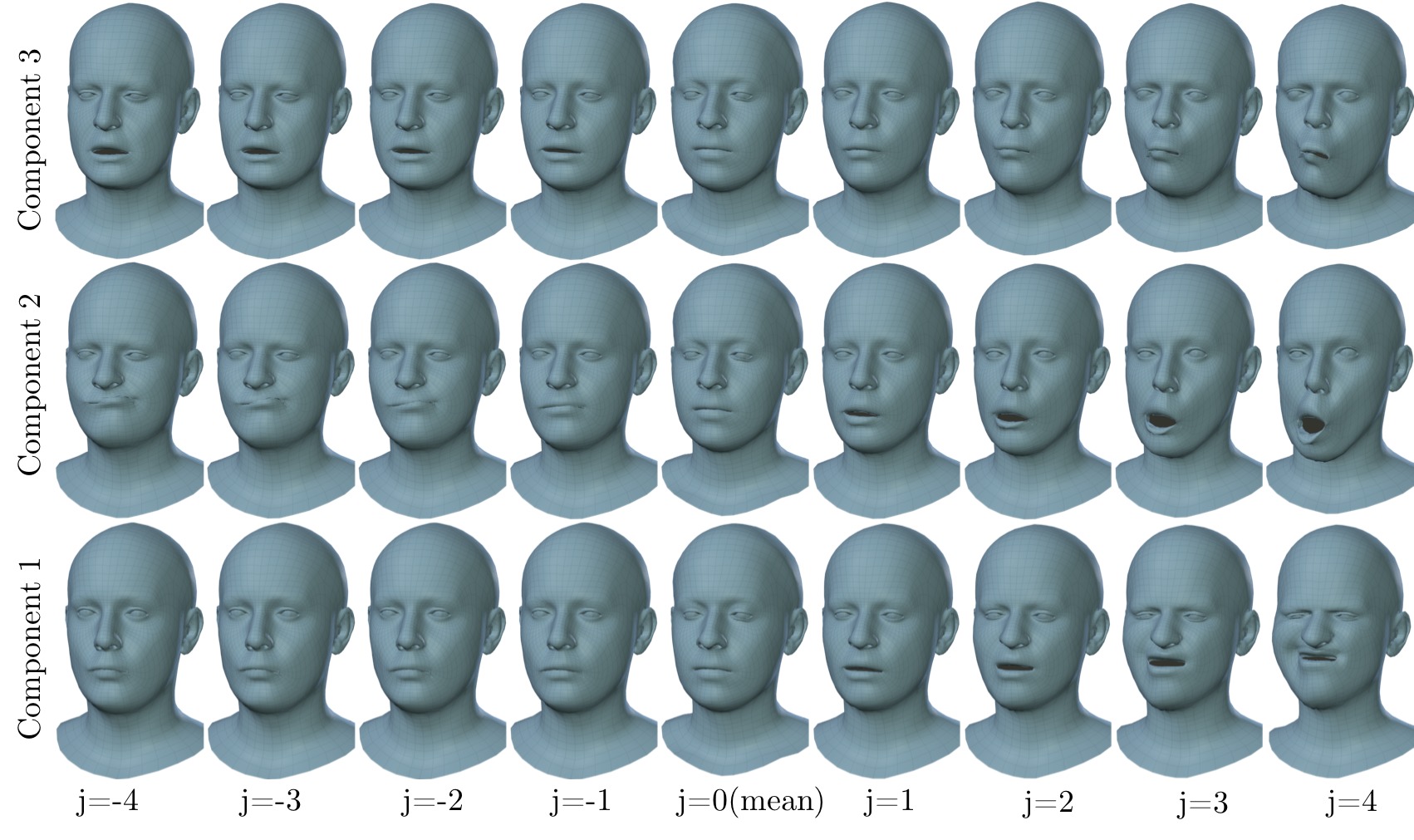}
\end{center}
\caption{Sampling from the latent space of the mesh autoencoder around the mean face $j=0$ along 3 different components.}
\label{fig:latent_space}
\end{figure}

\begin{figure}[t]
\begin{center}
\includegraphics[width=\linewidth]{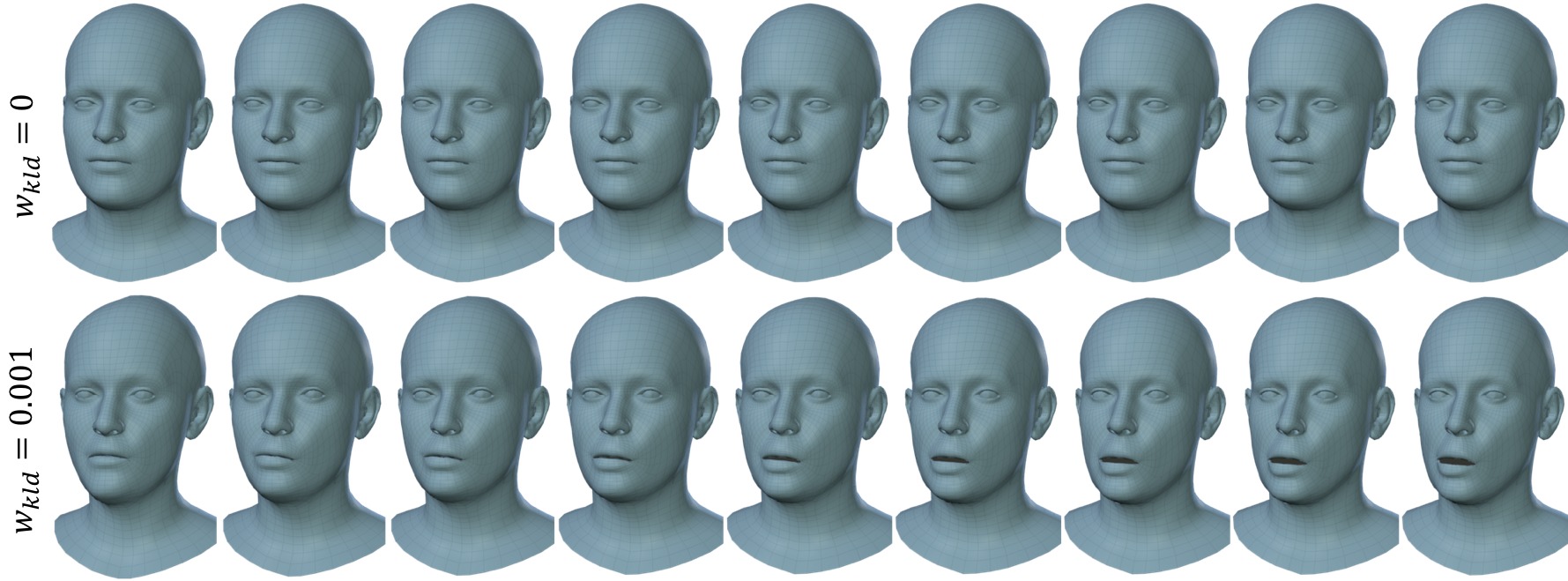}
\end{center}
\caption{Sampling using Gaussian noise with variational loss (bottom), and without(top). With $w_{kld}=0$, the latent representation might not have a Gaussian distribution. Hence, samples on top are not diverse.}
\label{fig:vae}
\end{figure}

\section{Experiments}
In this section, we evaluate the effectiveness of CoMA on an extreme facial expression dataset. We demonstrate that CoMA allows the synthesis of new expressive faces by sampling from the latent space in Section~\ref{sec:latent_sampling}, including the effect of adding variational loss. Following, we compare CoMA to the widely used PCA representation for reconstructing expressive 3D faces. For this, we evaluate in Section~\ref{sec:comparison_pca} the ability to reconstruct data similar to the training data (interpolation experiment), and the ability to reconstruct expressions not seen during training (extrapolation experiment). Finally, in Section~\ref{sec:deep_FLAME}, we show improved performance by replacing the expression space of state of the art face model, FLAME \cite{FLAME2017} with our autoencoder.

\subsection{Facial Expression Dataset}
Our dataset consists of 12 classes of extreme expressions from 12 different subjects. These expressions are complex and asymmetric. The expression sequences in our dataset are -- bareteeth, cheeks in, eyebrow, high smile, lips back, lips up, mouth down, mouth extreme, mouth middle, mouth side and mouth up. We show samples from our dataset and the number of frames of each captured sequence in the Supplementary Material.

The data is captured at 60fps with a multi-camera active stereo system
(3dMD LLC, Atlanta) with six stereo camera pairs, five speckle projectors, and
six color cameras.  Our dataset contains 20,466 3D Meshes, each with about 120,000 vertices. The data is pre-processed using a sequential mesh registration method~\cite{FLAME2017} to reduce the data dimensionality to 5023 vertices.

\subsection{Sampling the Latent Space}
\label{sec:latent_sampling}
Let $E$ be the encoder and $D$ be the decoder.
We first encode a face mesh from our test set in the latent space to obtain features $z=E(\mathcal{F})$. We then vary each of the components of the latent vector as $\tilde{z}_i = z_i + \epsilon$. We then use the decoder to transform the latent vector into a reconstructed mesh $\mathcal{\tilde{F}} =D(\tilde{z}) $. In Figure \ref{fig:latent_space}, we show a diversity of face meshes sampled from the latent space. Here, we extend or contract the latent vector along different dimensions by a factor of 0.3 such that $\tilde{z}_i = (1+0.3j) z_i $, where $j$ is the step. In Figure \ref{fig:latent_space}, $j \in [-4, 4]$, and the mean face $\mathcal{F}$ is shown in the middle of the row. More examples are shown in the Supplementary Material. \\

\qheading{Variational Convolutional Mesh Autoencoder.}
Although 3D faces can be sampled from our convolutional mesh autoencoder, the distribution of the latent space is not known. Therefore, sampling requires a mesh to be encoded in that space. In order to constrain the distribution of the latent space, we add a variational loss on our model. Let $E$ be the encoder, $D$ be the decoder, and $z$ be the latent representation of face $\mathcal{F}$. We minimize the loss,
\begin{equation}
	l = ||\mathcal{F} - D(z) ||_1 + w_{kld} KL(\mathcal{N}(0,1)||Q(z|\mathcal{F})),
\end{equation}
where $w_{kld}=0.001$ weights the $KL$ divergence loss. The first term minimizes the L1 reconstruction error, and the second term enforces a unit Gaussian prior $\mathcal{N}(0,1)$ with zero mean on the distribution of latent vectors $Q(z)$. This enforces the latent space to be a multivariate Gaussian. In Figure \ref{fig:vae}, we show visualizations by sampling faces from a Gaussian distribution on this space within $[-3\sigma, 3\sigma]$, where $\sigma=1$, is the variance of the Gaussian prior. We compare the visualizations by setting $w_{kld}=0$. We observe that $w_{kld}=0$ does not enforce any Gaussian prior on $P(z)$, and therefore sampling with Gaussian noise from this distribution results in limited diversity in face meshes. We show more examples in the Supplementary Material.

\subsection{Comparison with PCA Spaces}
\label{sec:comparison_pca}
Several face models use PCA space to represent identity and expression variations ~\cite{Tewari2017,FLAME2017,Amberg2008,Breidt2011,Yang2011}. We perform interpolation and extrapolation experiments to evaluate our performance. We use Scikit-learn~\cite{scikit-learn} to compute PCA coefficients. We consistently use an 8-dimensional latent space to encode the face mesh using both the PCA model and Mesh Autoencoder.

\begin{figure}[t]
  \begin{minipage}[b]{0.5\textwidth}
    \begin{center}
    \includegraphics[width=\textwidth]{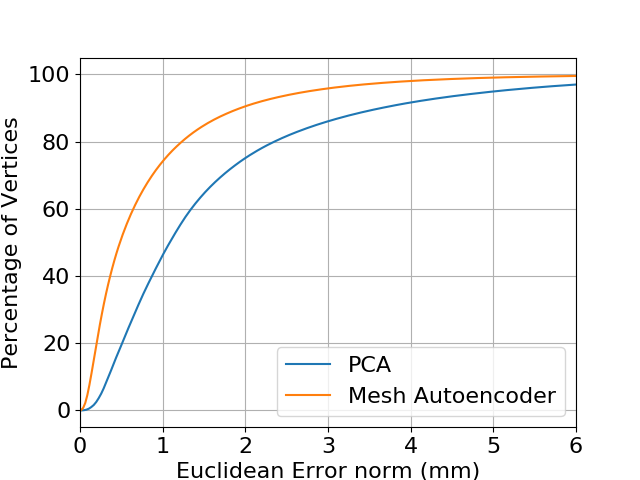} \\
    (a)
    \end{center}
  \end{minipage}
 \hfil
  \begin{minipage}[b]{0.5\textwidth}
    \begin{center}
    \includegraphics[width=\textwidth]{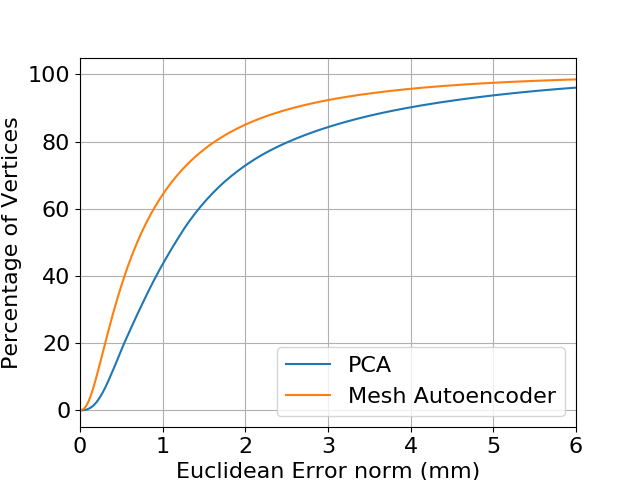} \\
    (b)
	\end{center}
\end{minipage}
\caption{Cumulative Euclidean error between PCA model and Mesh Autoencoder for Interpolation (a) and Extrapolation (b) experiments}
    \label{fig:cum_error1}
\end{figure}

\begin{figure}[t]
\begin{center}
\includegraphics[width=\linewidth]{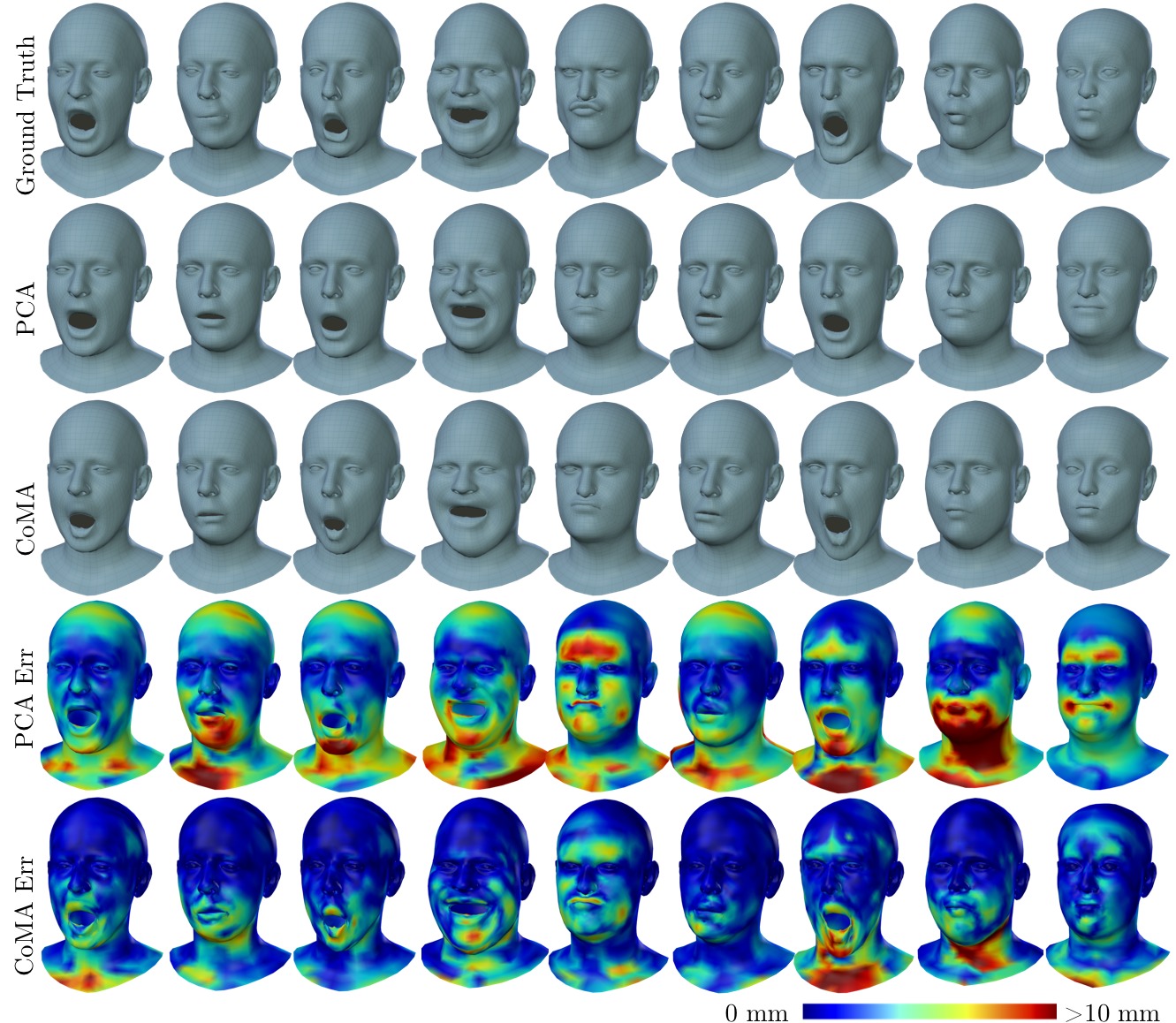}
\end{center}
\vspace{-4mm}
\caption{Comparison with PCA: Qualitative results for interpolation experiment}
\label{fig:interp}
\end{figure}

\qheading{Interpolation Experiment.}
In order to evaluate the interpolation capability of the autoencoder, we split the dataset in training and test samples with a ratio of 9:1. The test samples are obtained by picking consecutive frames of length 10  uniformly at random across the sequences. We train CoMA for 300 epochs and evaluate it on the test set. We use Euclidean distance for comparison with the PCA method. The mean error with standard deviation, and median errors are shown in Table \ref{tab:interp} for comparison.

\begin{table}
\begin{center}
\caption{Comparison with PCA: Interpolation experiments. Errors are in millimeters}
\begin{tabular}{l|cc|c}
     &  Mean Error      & Median Error & \# Parameters \\ \hline
PCA  & 1.639 $\pm$ 1.638 & 1.101& 120,552 \\
Mesh Autoencoder  & \textbf{0.845 $\pm$ 0.994} & \textbf{0.496}  & \textbf{33,856}\\
\end{tabular}
\label{tab:interp}
\end{center}
\end{table}

We observe that our reconstruction error is 50\% lower than PCA. At the same time, the number of parameters in CoMA is about 75\% fewer than the PCA model as shown in Table~\ref{tab:interp}. Visual inspection of our qualitative results in Figure \ref{fig:interp} shows that our reconstructions are more realistic and are effective in capturing extreme facial expressions. We also show the histogram of cumulative errors in Figure \ref{fig:cum_error1}a. We observe that our Mesh Autoencoder (CoMA) has about 72.6\% of the vertices within a Euclidean error of 1 mm, as compared to 47.3\% for the PCA model.

\begin{figure}[h]
\begin{center}
\includegraphics[width=\linewidth]{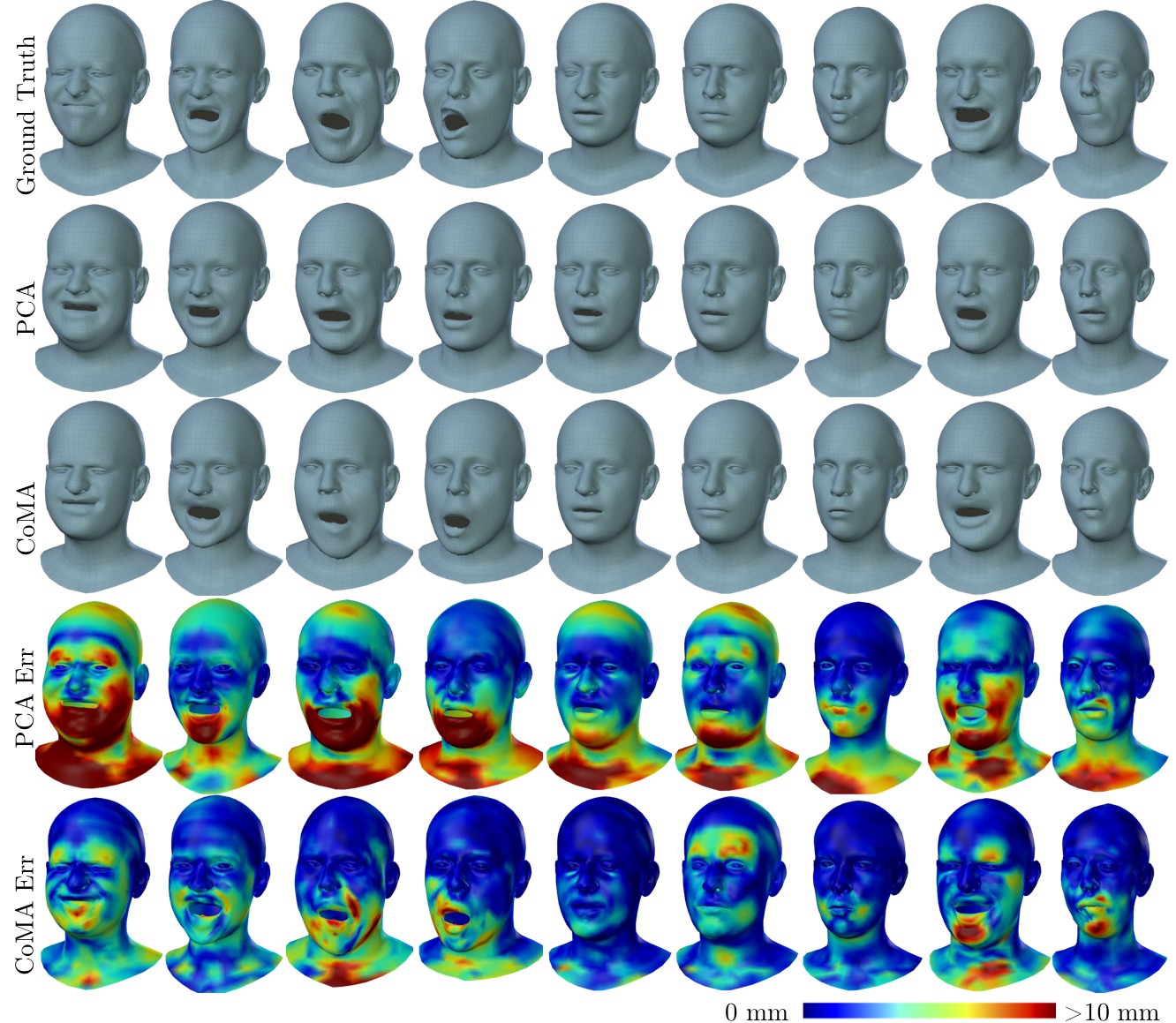}
\end{center}
\caption{Comparison with PCA: Qualitative results for extrapolation experiment}
\label{fig:extrap}
\end{figure}

\begin{table}[t]
\begin{center}
\caption{Comparison with PCA: Extrapolation experiment. Errors are in millimeters.}
\begin{tabular}{l|cc|cc|cc}
&  \multicolumn{2}{|c}{Mesh Autoencoder} & \multicolumn{2}{|c}{PCA} & \multicolumn{2}{|c}{FLAME \cite{FLAME2017}} \\
Sequence   &  Mean Error & Median & Mean Error & Median & Mean Error & Median \\ \hline
bareteeth & \textbf{1.376$\pm$1.536} & \textbf{0.856}  &1.957$\pm$1.888 & 1.335 & 2.002$\pm$1.456 & 1.606 \\
cheeks in & \textbf{1.288$\pm$1.501} & \textbf{0.794}  &1.854$\pm$1.906 & 1.179 & 2.011$\pm$1.468 & 1.609 \\
eyebrow &  \textbf{1.053$\pm$1.088} & \textbf{0.706} &1.609$\pm$1.535 & 1.090 & 1.862$\pm$1.342 & 1.516 \\
high smile & \textbf{1.205$\pm$1.252} & \textbf{0.772} & 1.841$\pm$1.831 & 1.246 & 1.960$\pm$1.370 & 1.625 \\
lips back & \textbf{1.193$\pm$1.476} & \textbf{0.708} & 1.842$\pm$1.947 & 1.198 & 2.047$\pm$1.485 & 1.639 \\
lips up & \textbf{1.081$\pm$1.192} & \textbf{0.656} & 1.788$\pm$1.764 & 1.216 & 1.983$\pm$1.427 & 1.616 \\
mouth down & \textbf{1.050$\pm$1.183} & \textbf{0.654} & 1.618$\pm$1.594 & 1.105 & 2.029$\pm$1.454 & 1.651 \\
mouth extreme &\textbf{1.336$\pm$1.820} & \textbf{0.738}  & 2.011$\pm$2.405 & 1.224 & 2.028$\pm$1.464 & 1.613 \\
mouth middle & \textbf{1.017$\pm$1.192} & \textbf{0.610}  &1.697$\pm$1.715 & 1.133 & 2.043$\pm$1.496 & 1.620 \\
mouth open & \textbf{0.961$\pm$1.127} & \textbf{0.583}  &1.612$\pm$1.728 & 1.060 & 1.894$\pm$1.422 & 1.544 \\
mouth side &  \textbf{1.264$\pm$1.611} & \textbf{0.730}  &1.894$\pm$2.274 & 1.132& 2.090$\pm$1.510 & 1.659 \\
mouth up & \textbf{1.097$\pm$1.212} & \textbf{0.683}  &1.710$\pm$1.680 & 1.159 & 2.067$\pm$1.485 & 1.680
\end{tabular}
\vspace{-4mm}
\label{tab:extrap}
\end{center}
\end{table}

\qheading{Extrapolation Experiment.} To measure generalization of our model, we compare the performance of CoMA with the PCA model and FLAME \cite{FLAME2017}. For comparison, we train the expression model of FLAME on our dataset. The FLAME reconstructions are obtained with latent vector size of 16 with 8 components each for encoding identity and expression. The latent vectors encoded using the PCA model and Mesh autoencoder have a size of 8.

To evaluate generalization capability of our model, we reconstruct the expressions that are completely unseen by our model. We perform 12 different experiments for evaluation. For each experiment, we split our dataset by completely excluding one expression set from all the subjects of the dataset. We test our Mesh Autoencoder on the excluded expression. 
We compare the performance of our model with PCA and FLAME using the Euclidean distance (mean, standard deviation, median). We perform 12 fold cross validation, one for each expression as shown in Table \ref{tab:extrap}.
In Table \ref{tab:extrap}, we also show that our model performs better than PCA and FLAME \cite{FLAME2017} on all expression sequences. We show the qualitative results in Figure \ref{fig:extrap}.
We show the cumulative Euclidean error histogram in Figure \ref{fig:cum_error1}b. For a 1 mm accuracy, Mesh Autoencoder captures 63.8\% of the vertices while the PCA model captures 45\%.

\subsection{DeepFLAME}
\label{sec:deep_FLAME}
FLAME~\cite{FLAME2017} is a state of the art model for face representation that combines linear blendskinning for head and jaw motion with linear PCA spaces to represent identity and expression shape variations. To improve the reconstruction error of FLAME, we replace the PCA expression space of FLAME with our autoencoder, and refer to the new model as DeepFLAME. We compare the performance of DeepFLAME with FLAME by varying the size of the latent vector for encoding. Head rotations are factored out for comparison since they are well modeled by linear blendskinning in FLAME, and we consider only the expression space. The reconstruction accuracy is measured using Euclidean distance metric. We show the comparisons in Table \ref{tab:ablexp}. The median reconstruction of DeepFLAME is lower for all chosen latent space dimensions, while the mean reconstruction error is lower for up to 12 latent variables. This shows that DeepFLAME provides a more compact face representation; i.e., captures more shape variation with fewer latent variables.


\begin{table}[t]
\begin{center}
\caption{Comparison of FLAME and DeepFLAME. DeepFLAME is obtained by replacing expression model of FLAME with CoMA. All errors are in millimeters.}
\begin{tabular}{l|cc|cc}
&  \multicolumn{2}{|c}{DeepFLAME} & \multicolumn{2}{|c}{FLAME \cite{FLAME2017}}  \\
\#dim of $z$   &  Mean Error & Median & Mean Error & Median  \\ \hline
2 & \textbf{0.610$\pm$0.851} & \textbf{0.317} & 0.668$\pm$0.876 & 0.371 \\
4 &  \textbf{0.509$\pm$0.746} & \textbf{0.235} &0.589$\pm$0.803 & 0.305 \\
6 &  \textbf{0.464$\pm$0.711} & \textbf{0.196} &0.525$\pm$0.743 & 0.252 \\
8 &  \textbf{0.432$\pm$0.681} & \textbf{0.169} &0.477$\pm$0.691 & 0.217 \\
10 &  \textbf{0.421$\pm$0.664} & \textbf{0.162} &0.439$\pm$0.655 & 0.193 \\
12 &  \textbf{0.388$\pm$0.630} & \textbf{0.139} &0.403$\pm$0.604 & 0.172 \\
14 &  0.371$\pm$0.605 & \textbf{0.128} &\textbf{0.371$\pm$0.567} & 0.152 \\
16 &  0.372$\pm$0.611 & \textbf{0.125} &\textbf{0.351$\pm$0.543} & 0.139
\end{tabular}
\label{tab:ablexp}
\end{center}
\end{table}

\subsection{Discussion}

The focus of CoMA is to model facial shape for reconstruction applications. The Laplace-Beltrami operator (LBo) describes the intrinsic surface geometry and is invariant under isometric surface deformations. This isometry invariance of the LBo is beneficial for shape matching and registration. Since changes in facial expression are near isometric deformations~\cite[Section 13.3]{Bronstein2008}, applying LBo to expressive faces would result in a loss of most expression-related shape variations, making it infeasible to model such variations. The graph Laplacian used by CoMA in contrast to the LBo is not isometry invariant.

While we evaluate CoMA on face shapes, it is applicable to any class of objects. Similar to existing statistical models however, it requires all meshes in dense vertex correspondence; i.e. all meshes need to share the same topology. A future research direction is to directly learn a 3D face representation from raw 3D face scans or 2D images without requiring vertex correspondence.

As is also true for other deep learning based models, the performance of CoMA could further improve with more training data. The amount of existing 3D face data however is very limited. The data scarcity especially limits our expression model to outperform existing models for higher latent space dimensions ($> 12$ see Table~\ref{tab:ablexp}). We predict superior quality on larger datasets and plan to evaluate CoMA on significantly more data in the future.

As CoMA is an end-to-end trained model, it could also be combined with some existing image convolutional network to regress the 3D face shape from 2D images. We will explore this in future work. 



\section{Conclusion}

We have introduced CoMA, a new representation for 3D faces of varying shape and expression. We designed CoMA as a hierarchical, multi-scale representation to capture global and local shape and expression variations of multiple scales. To do so, we introduce novel sampling operations and combine these with fast graph convolutions in an autoencoder network. The locally invariant filters, shared across the mesh surface, significantly reduce the number of filter parameters in the network, and the non-linear activation functions capture extreme facial expressions. We evaluated CoMA on a dataset of extreme 3D facial expressions that we will make publicly available for research purposes along with the trained model. We showed that CoMA significantly outperforms state-of-the-art models in 3D face reconstruction applications while using $75\%$ fewer model parameters. CoMA outperforms the linear PCA model by $50\%$ on interpolation experiments and generalizes better on completely unseen facial expressions. We further demonstrated that CoMA in a variational setting allows us to synthesize new expressive faces by sampling the latent space. \\

\section*{Acknowledgement}
We thank Tsvetelina Alexiadis and Jorge M\'{a}rquez for data aquisition. We thank Haven Feng for rendering the figures. We acknowledge the advice of Stefanie Wuhrer on mesh convolutions. We are grateful to Georgios Pavlakos, Despoina Paschalidou and Sergi Pujades for helping us with several revisions of the paper.
\bibliographystyle{splncs04}
\bibliography{bibliography}

\section*{S. Details of the Dataset}
We capture 3D sequences of 12 subjects of different age groups, each of whom perform 12 different expressions. These expressions are chosen such that they are extreme causing a lot of facial tissue deformation. We also make sure that no two expressions are correlated 
with each other. The number of frames for each expression is listed in Table \ref{tab:dataset_len}.

\begin{table}[t]
\begin{center}
\begin{tabular}{l|c}
\textbf{Sequence} &\textbf{\# Frames}\\
\hline
bareteeth & 1946 \\
cheeks in & 1396 \\
eyebrow & 2283 \\
high smile & 1878 \\
lips back & 1694 \\
lips up &1511 \\
mouth down & 2363 \\
mouth extreme & 793 \\
mouth middle & 1997 \\
mouth open & 674 \\
mouth side & 1778 \\
mouth up & 2153
\end{tabular}
\end{center}
\caption{Length of different expression sequences}
\label{tab:dataset_len}
\end{table}

\begin{figure}[t]
\begin{center}
\includegraphics[width=\linewidth]{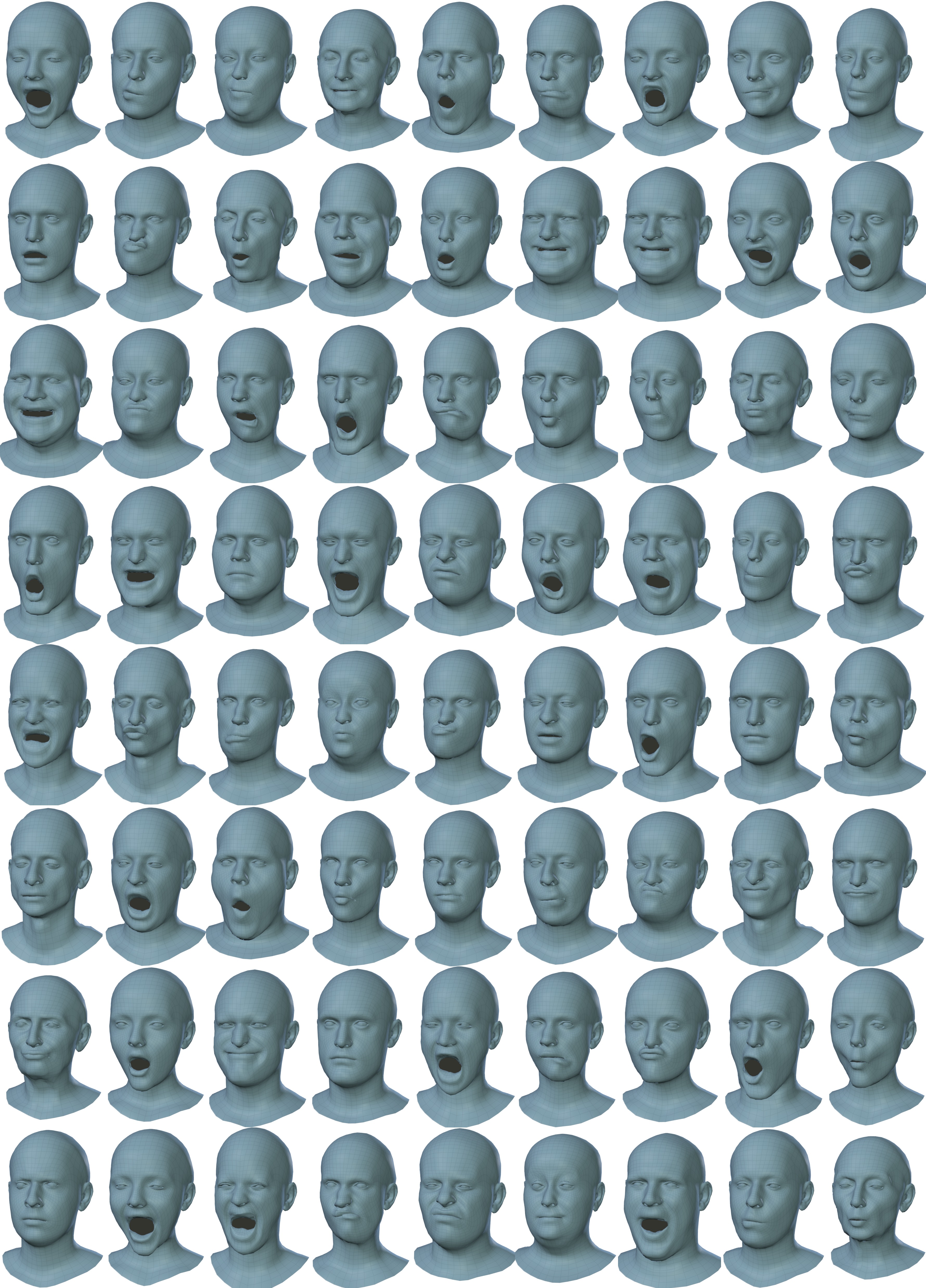}
\end{center}
\vspace{-6mm}
\caption{Samples from the dataset}
\label{fig:data_samples}
\end{figure}

\begin{figure}[t]
\begin{center}
\includegraphics[width=\linewidth]{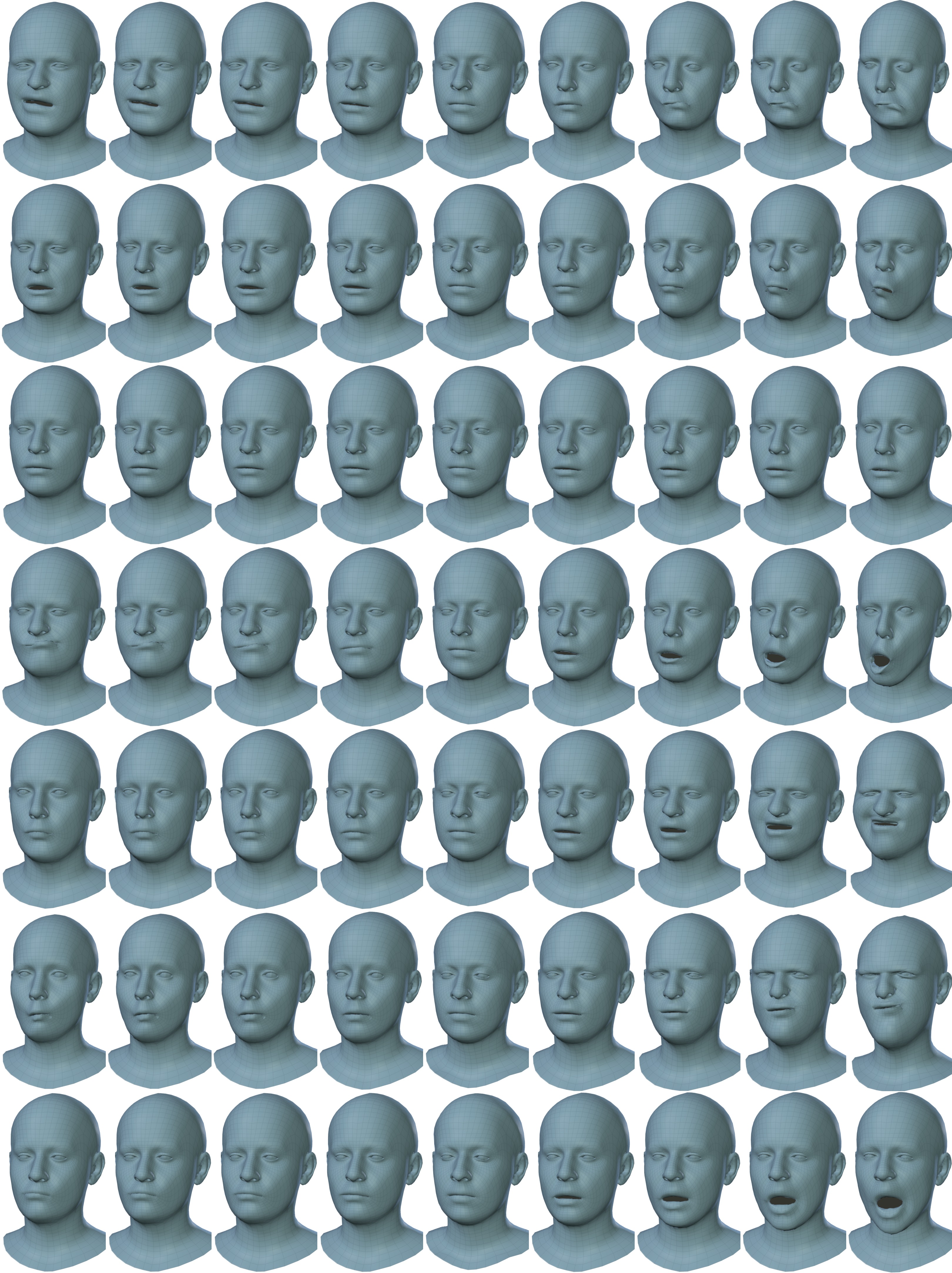}
\end{center}
\vspace{-6mm}
\caption{Sampling from latent space of CoMA, each row is sampled along a particular dimension.}
\label{fig:coma_latent}
\end{figure}

\begin{figure}[t]
\begin{center}
\includegraphics[width=\linewidth]{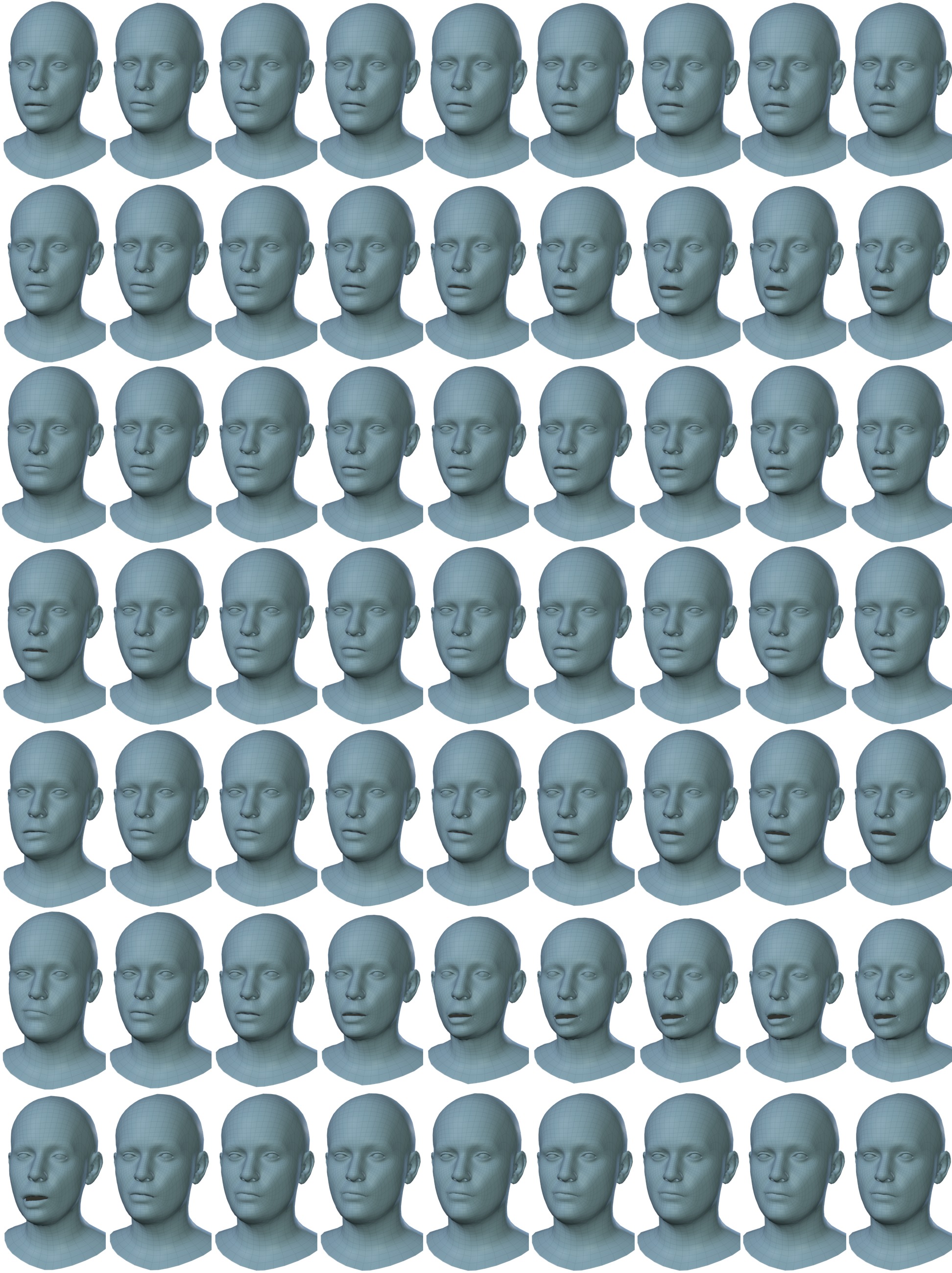}
\end{center}
\vspace{-6mm}
\caption{Sampling from latent space of Variational CoMA, each row is sampled along a particular dimension.}
\label{fig:vae_coma_latent}
\end{figure}

\end{document}